\documentclass{article} % For LaTeX2e
\pdfoutput=1 % force pdflatex on arXiv
\PassOptionsToPackage{table}{xcolor}
\usepackage[preprint]{colm2026_conference}

\usepackage{microtype}
\usepackage{graphicx}
\graphicspath{{figs/}}
\usepackage{amsmath}
\usepackage{amssymb}
\usepackage{bm}
\usepackage{booktabs}
\usepackage{tabularx}
\usepackage{multirow}
\usepackage{placeins}    % \FloatBarrier keeps tables within their section
\usepackage{xcolor}
\usepackage{marvosym}    % \Letter (envelope) for corresponding-author mark
\usepackage[most]{tcolorbox}
\usepackage{sectsty}     % recolor section headings
\usepackage{hyperref}
\usepackage{url}
\usepackage{cleveref}  % must be loaded after hyperref and amsmath
\usepackage{lineno}      % defines \@LN@col etc.; safe even when line numbers are off (preprint)

\definecolor{ourcolor}{rgb}{0.85,0.93,1.00}
\definecolor{deltaup}{rgb}{0.00,0.55,0.00}
\definecolor{deltadown}{rgb}{0.75,0.00,0.00}
\newcommand{\du}[1]{{\scriptsize\textcolor{deltaup}{$\uparrow$#1}}}
\newcommand{\dd}[1]{{\scriptsize\textcolor{deltadown}{$\downarrow$#1}}}

\definecolor{uwblue}{rgb}{0.20,0.50,0.80}
\hypersetup{colorlinks=true, allcolors=uwblue}
\sectionfont{\color{uwblue}}
\subsectionfont{\color{uwblue}}
\subsubsectionfont{\color{uwblue}}
\newcommand{\corr}{\textsuperscript{\Letter}}  % corresponding-author mark

\newtcolorbox{abstractbox}{%
  enhanced, breakable=false,
  colback=white, colframe=uwblue,
  boxrule=0.8pt, arc=6pt,
  left=16pt, right=16pt, top=12pt, bottom=14pt,
  width=\textwidth, before skip=8pt, after skip=16pt,
}

\fancypagestyle{titlepage}{%
  \fancyhf{}%
  \renewcommand{\headrulewidth}{0pt}%
  \fancyfoot[C]{\thepage}%
}

\begin{document}

\ifcolmsubmission
\linenumbers
\fi

\thispagestyle{titlepage}
\vspace*{-80pt}

% ---- Institution logo (top-left); the thick rule below it is the first line ----
\noindent\includegraphics[height=40pt]{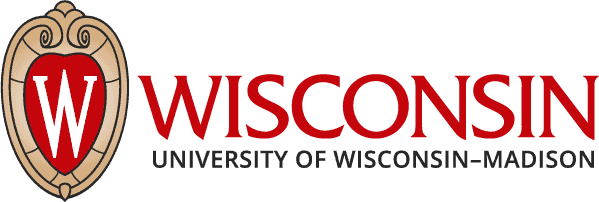}\par
\vspace{-4pt}
\noindent{\color{uwblue}\rule{\textwidth}{2.5pt}}
\vspace{4pt}                 % equal gap above title

% ---- Title (equal spacing to the rules above and below) ----
\centerline{\LARGE\bfseries Test-Time Training with Next-Token Prediction}
\vspace{4pt}                 % equal gap below title
\noindent{\color{uwblue}\rule{\textwidth}{0.5pt}}
\vspace{3pt}                 % small gap: authors sit close to the thin rule

% ---- Authors (corresponding author = last author) + affiliation ----
{\centering
  {\large Xuan Ouyang\textsuperscript{*}\quad Zefan Cai\textsuperscript{*}\quad Junjie Hu\corr\par}
  \vspace{1pt}
  {\small \textsuperscript{*}Equal Contribution\quad\textsuperscript{\Letter}\,Corresponding Author\par}
  \vspace{1pt}
  {\normalsize University of Wisconsin--Madison\par}
  \par}
\vspace{4pt}

% ---- Boxed abstract ----
\begin{abstractbox}
{\centering {\large\bfseries\color{uwblue} Abstract}\par}
\vspace{6pt}
\noindent
Next-token prediction is the self-supervised signal that trains language
models, and every observed prompt token provides the same signal at test
time. We study whether this signal can define the inner-loop objective for
test-time training (TTT) in pretrained long-context language models. Many TTT architectures require models to be trained with test-time adaptation in mind, limiting their direct applicability to released LLM checkpoints. While recent in-place TTT methods make fast-weight adaptation possible for pretrained LLMs without redesigning the backbone, they leave a central question unresolved: what should each test-time write store? Existing recipes train the fast weight to match a learned local value proxy but they are not directly tied to the self-supervised next-token prediction signal. We introduce Test-Time Training with
Next-Token Prediction (TTT-NTP), a drop-in fast-weight adaptation method
for pretrained LLMs that instead supervises updates using the model's own
next contextual hidden state. This makes each local write follow the same
causal computation that supports next-token prediction: the value target is a pointwise
linear projection of a single next-position contextual state. On RULER Full-13, averaged over 4k to 32k contexts, TTT-NTP is the only method that consistently improves the released backbone across four models spanning three families and a 0.6--8B size range, by 3.9 points on Llama-3.1-8B, 3.0 on Mistral-7B-v0.3, 4.1 on Qwen3-4B, and 2.9 on Qwen3-0.6B. On the real-world LongBench-v2 long-document QA benchmark, TTT-NTP improves over the base model by 5.6 points on Llama-3.1-8B and 3.7 on Mistral-7B-v0.3, while preserving commonsense and knowledge performance. Our code is publicly available at \url{https://github.com/yancyou/TTT-NTP}.
\end{abstractbox}

\section{Introduction}
\label{sec:intro}

Language models are trained by next-token prediction (NTP): given a prefix,
predict the next observed token. The same self-supervised signal is present at
test time for every token in a prompt. Before the model generates a response,
the prompt itself provides many prefix--next-token pairs that can adapt the
model to the current document, topic, style, or retrieval problem. This makes
NTP a natural training signal for test-time training (TTT) in language models.

The need for such adaptation is clearest in long-context settings. Modern
large language models (LLMs) support long context windows
\citep{grattafiori2024llama}, and positional-extension methods
\citep{su2024roformer,peng2024yarn,ding2024longrope} and long-document
continual pretraining \citep{chen2024longlora,fu2024data} have substantially
increased their nominal context lengths. Yet long-context benchmarks show
that models still fail to use information that lies inside the available
window \citep{liu2024lost,hsieh2024ruler,bai2024longbench}. These failures
are not simply failures of context capacity: the evidence is present, but
the model's computation does not carry it forward reliably enough to affect
the next-token distribution.

Test-time training offers a direct mechanism for changing how prompt
information is represented during inference. Instead of relying only on
attention over a fixed key--value cache, the model writes observations into
\emph{fast weights}
\citep{schmidhuber1992learning,ba2016using,ramsauer2020hopfield} that are
read by later tokens. This dynamic-memory view is closely related to linear
and recurrent sequence models
\citep{katharopoulos2020transformers,gu2023mamba,dao2024transformers,sun2023retentive,peng2023rwkv,yang2024parallelizing}
and to broader test-time memory systems \citep{behrouz2026titans}. Recent
TTT methods make fast-weight adaptation practical for pretrained LLMs by
placing fast weights at selected MLP down-projections and accumulating
chunk-parallel rank-one writes \citep{feng2026place}. However, the
objective that drives these writes remains misaligned with the model's own
prediction objective. Existing recipes are trained with an outer
language-modeling loss, but the inner loop writes a learned local activation
proxy into the fast weight.

We introduce \emph{Test-Time Training with Next-Token Prediction} (TTT-NTP).
TTT-NTP aligns the fast-weight inner loop with the same self-supervised
prediction problem used to train the backbone. Directly taking full
next-token cross-entropy gradients through every adapted layer at test time
would be expensive and would destroy the chunk-parallel structure that makes
in-place TTT attractive. TTT-NTP therefore uses a layer-local form of the
NTP signal: at an adapted MLP block, the key is the current gated MLP
activation, and the value target is the same layer's contextual hidden state
at the next position. This state is produced by the causal forward pass on
the observed next token. Because it lies on the representation trajectory
used to form later next-token distributions, it provides a dense,
model-native target without aggregating over a neighborhood of positions.

During continual pretraining, TTT-NTP learns this chunk-parallel causal
write. At inference time, the same next-position rule yields a closed-form
fast-weight write computed from cached gated activations, which we apply
before decoding. On RULER Full-13, averaged over 4k to 32k contexts, TTT-NTP is the only method that consistently improves the released backbone across four models from three families spanning $0.6$--$8$B, and on the real-world LongBench-v2 long-document QA benchmark it again attains the best overall accuracy on both Llama-3.1-8B and Mistral-7B-v0.3. Under matched data, compute,
fast-weight placement, and chunk size, the prior
in-place TTT method does not improve over the released backbone on RULER, supporting
the importance of the next-token-prediction-aligned target. General
capability on ARC \citep{clark2018think}, PIQA \citep{bisk2020piqa}, MMLU
\citep{hendrycks2020measuring}, and HellaSwag \citep{zellers2019hellaswag}
remains comparable in aggregate.

\paragraph{Contributions.}
\begin{itemize}
  \item We formulate TTT-NTP, a test-time-training framework that aligns fast-weight writes in pretrained LLMs with next-token prediction, instantiated with a layer-local next-position target that pairs each current MLP activation with the next contextual hidden state produced by the causal forward pass.
  \item Across four backbones from three families spanning $0.6$--$8$B, TTT-NTP is the only method that consistently improves RULER Full-13 over the released model, by $3.9$ points on Llama-3.1-8B, $3.0$ on Mistral-7B-v0.3, $4.1$ on Qwen3-4B, and $2.9$ on Qwen3-0.6B, while keeping aggregate commonsense and knowledge performance comparable.
  \item On the real-world LongBench-v2 long-document QA benchmark, TTT-NTP attains the best overall accuracy on both evaluated backbones, improving over Base by $5.6$ points on Llama-3.1-8B and $3.7$ on Mistral-7B-v0.3.
  \item A target ablation isolates the supervisory signal from the rank-one machinery: under matched placement, update mechanics, data, and compute, next-position supervision outperforms Past-5, Next-5, and Bi-dir-5 convolutional targets by at least five RULER points at every evaluated length.
\end{itemize}

\section{Related Work}
\label{sec:related}

\subsection{Test-Time Training for Language Models}
\label{sec:related:ttt}

Test-time training adapts a model on each test input using a self-supervised
objective \citep{sun2020test}. In later architectural forms, the hidden state
of a recurrent layer is itself a set of fast weights updated by an online
learner \citep{sun2024learning}. This view connects naturally to fast-weight
programmers \citep{schmidhuber1992learning,ba2016using}, the fast-weight
interpretation of linear transformers \citep{schlag2021lineartransformerssecretlyfast}, modern
Hopfield networks \citep{ramsauer2020hopfield}, and linear-state sequence
models such as linear attention \citep{katharopoulos2020transformers}, RetNet
\citep{sun2023retentive}, RWKV \citep{peng2023rwkv}, DeltaNet
\citep{yang2024parallelizing,yang2025gated}, and state-space models
\citep{gu2023mamba,dao2024transformers}. All maintain a compact state that is
written by earlier tokens and read by later tokens.

Recent work makes TTT practical for long-context language models through
large-chunk or in-place fast-weight updates
\citep{zhang2025test,tandon2025end,behrouz2026titans}; most directly,
In-Place TTT \citep{feng2026place} updates existing MLP down-projections with
chunk-parallel rank-one writes. With the write mechanism and placement largely fixed by these methods, the
supervisory target is the open design axis. Prior in-place recipes use the current representation or a
locally constructed value proxy, often built from neighboring activations by a small auxiliary network \citep{feng2026place,behrouz2026titans}. TTT-NTP changes this supervision target. Instead of asking the fast weight to reconstruct a local proxy, we use the next-position contextual hidden state at the same layer: a model-native representation produced by the causal forward pass after observing the next prompt token.

\subsection{Next-Token Prediction and Long Context}
\label{sec:related:longctx}

Next-token prediction is the standard objective for autoregressive language
modeling, but optimizing the objective on long documents does not by itself
ensure reliable long-context utilization. One line of work extends effective
context windows through rotary position embeddings \citep{su2024roformer},
positional interpolation \citep{chen2023extending},
frequency-aware rescaling \citep{peng2024yarn,ding2024longrope}, or ALiBi
\citep{press2021train}, often combined with efficient attention kernels
\citep{dao2022flashattentionfastmemoryefficientexact}. A second line modifies attention with local or sparse
patterns \citep{beltagy2020longformerlongdocumenttransformer,zaheer2021bigbirdtransformerslonger}, or replaces
attention with recurrent or linear-state mechanisms
\citep{katharopoulos2020transformers,gu2023mamba,dao2024transformers,sun2023retentive,peng2023rwkv,yang2024parallelizing}.
A third performs continual pretraining on long documents with tuned data
mixtures and long-context recipes
\citep{chen2024longlora,fu2024data,grattafiori2024llama,yang2025qwen3}.

Despite these advances, evaluation suites such as RULER and LongBench show
substantial degradation inside the nominal context window
\citep{hsieh2024ruler,bai2024longbench,liu2024lost}. The relevant evidence is
often in the prompt, but the model fails to keep it available for the future
next-token decisions that need it. TTT-NTP is complementary to context-window
extension and data selection \citep{wang2026opus}: it keeps the pretrained backbone and context window fixed, then uses the prompt's own next-token pairs to update a compact fast-weight memory.

\subsection{Predictive Targets in Representation Space}
\label{sec:related:predictive}

TTT-NTP uses an NTP-aligned signal but implements it through a
representation-space target. This follows a broader lesson from
self-supervised learning: predictive targets need not be raw observations.
BYOL predicts another view's embedding \citep{grill2020bootstrap}; I-JEPA predicts
masked image regions in latent space \citep{assran2023self}; and DINOv2 uses
large-scale latent-space distillation for visual representation learning
\citep{oquab2024dinov}. Masked autoencoders \citep{he2022masked} reconstruct
through a decoder bottleneck, sharing the idea that an intermediate
representation can carry richer and more stable supervision than the input
alone.

For LLM TTT, directly writing vocabulary-level targets into a layer-local fast
weight is poorly matched to the MLP down-projection where the fast weight
lives. The next contextual hidden state is a better interface: it is dense,
same-dimensional as the MLP output stream, and causally determined by the next
observed token. Unlike hidden-state distillation losses that only shape static
parameters during training, TTT-NTP uses the next hidden state as the value
target for a fast weight that is rewritten on each prompt.

\section{Method}
\label{sec:method}

TTT-NTP places a small \emph{fast weight} inside selected MLP blocks and updates it from a next-token signal while reading the input. During continual pretraining the fast-weight write is learned in a chunk-parallel causal form (\S\ref{sec:method:prelim}–\ref{sec:method:chunk}); at inference the same target yields a closed-form prompt write applied before decoding (\S\ref{sec:method:cf}).

\subsection{From TTT to Next-Token Prediction}
\label{sec:method:prelim}

Test-time training updates a small subset of model parameters on the test
input itself using a self-supervised loss \citep{sun2020test,sun2024learning}. For
an autoregressive language model, the most natural self-supervised signal is
already present: for every observed prompt token $x_{t+1}$, the prefix
$x_{1:t}$ defines the next-token prediction loss
\begin{equation}
\mathcal{L}_{\mathrm{NTP}}(x_{1:t+1})
=
-\log p_{\theta}(x_{t+1}\mid x_{1:t}).
\label{eq:ntp-loss}
\end{equation}
A naive TTT implementation could take gradient steps on
\cref{eq:ntp-loss} while reading the prompt. However, this is computationally expensive for large decoders, hard to parallelize, and touches a broad portion of the network. We therefore confine the adaptation to a compact \emph{fast weight} $\bm{W}$ inside each adapted MLP block. Given a key $\bm{k}_t$ and a value target $\bm{v}_t$, we take a single gradient step that aligns the key's linear projection $\bm{W}\bm{k}_t$ with the target,
\begin{equation}
\bm{W} \leftarrow \bm{W} - \eta \nabla_{\bm{W}} \mathcal{L}\bigl(\bm{W}\bm{k}_t, \bm{v}_t\bigr),
\qquad
\mathcal{L}\bigl(\bm{W}\bm{k}_t, \bm{v}_t\bigr) = -\bigl\langle \bm{W}\bm{k}_t, \bm{v}_t \bigr\rangle .
\label{eq:ttt-update}
\end{equation}

% Specifically, in a forward pass of each token $x_t$, its key vector $\bm{k}_t$ is linearly mapped through $\bm{W}$ to a target value vector $\bm{v}_t$, and $\bm{W}$ is optimized by a squared L2 loss that minimizes the Euclidean distance between $\bm{W}\bm{k}_t$ and $\bm{v}_t$: %typically by linearly mapping a key $\bm{k}_{t}$ via $\bm{W}$ to a value target %\jh{The $\mathcal{L}$ function in Eq.~\ref{eq:ttt-update} is not clearly defined here. It's a bit unclear how it is related to $\mathcal{L}_{NTP}$ above. Another point: when you say ``fast weight W inside the model'', are you trying to be more generic? Can we be a bit more explicit and say inside ``the MLP blocks of the model''?}
% \begin{align}
%     \bm{W}
%     \leftarrow
%     \bm{W}
%     - \eta\,\nabla_{\bm{W}}\,
%     \mathcal{L}_2\!\bigl(\bm{W}\bm{k}_{t},\,\bm{v}_{t}\bigr).
%     \label{eq:ttt-update}
% \end{align}
% \begin{equation}
% \bm{W}_{t}
% \leftarrow
% \bm{W}_{t-1}
% - \eta\,\nabla_{\bm{W}}\,
% \mathcal{L}\!\bigl(\bm{W}_{t-1}\bm{k}_{t},\,\bm{v}_{t}\bigr).
% \label{eq:ttt-update}
% \end{equation}
In practice we accumulate these rank-one writes over chunks of tokens for parallelism, rather than applying a separate update at every position (details in \S\ref{sec:method:chunk}).

The remaining design choice is the target $\bm{v}_t$, which determines what the fast weight stores. TTT-NTP uses the model's own next-position contextual state: when $x_{t+1}$ enters the causal forward pass, it produces—at the adapted layer—the hidden state the model would propagate next, and we write that state into the fast weight rather than a separately learned target. Because this state lies on the same causal trajectory the model uses to form later next-token distributions, the write nudges the MLP output along the model's own predictive path, and it is available densely at every position from a single forward pass.

% The central design question is therefore the choice of the local target $\bm{v}_{t}$. TTT-NTP chooses a target induced by next-token prediction. When the next token $x_{t+1}$ enters the causal forward pass, it produces a contextual state at every layer. This next-position state is the representation the model propagates after observing the next token, and it is the state from which later next-token predictions continue. We use this representation as a layer-local proxy for the NTP objective. Rather than training the MLP fast weight to predict vocabulary logits directly, we train it to reconstruct the hidden state that the network produces after observing the next token. Because this hidden state is generated by the same computation that ultimately supports next-token prediction, matching it provides an efficient local approximation to the global NTP learning signal.
%We use it as a layer-local surrogate for the NTP objective: instead of training an MLP fast weight to predict vocabulary logits, we train it to write the next contextual state that the NTP computation creates inside the network.

\subsection{TTT-NTP Fast-Weight Write}
\label{sec:method:ntp}

\begin{figure*}[t]
  \centering
  \includegraphics[width=\textwidth]{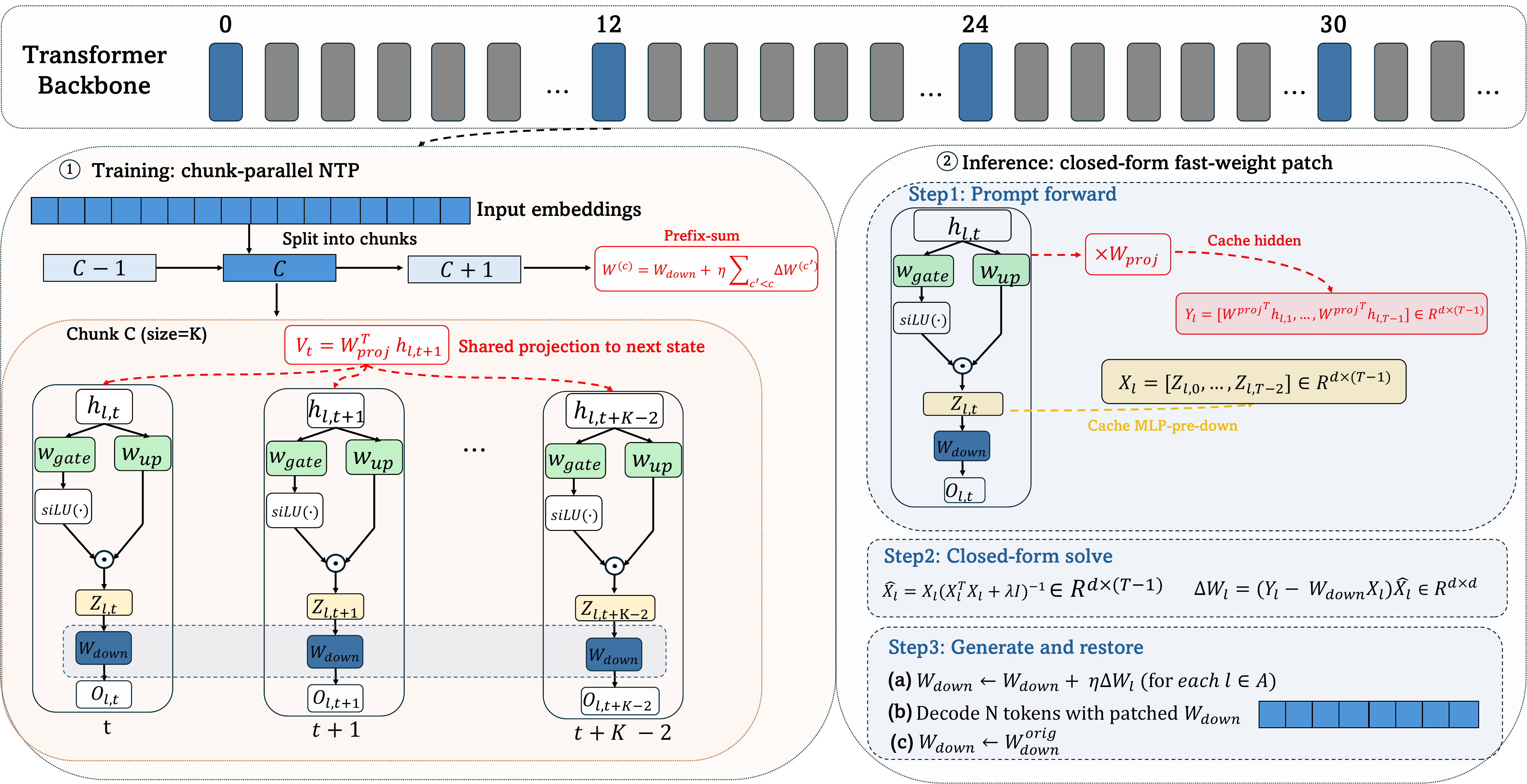}
  \caption{\textbf{Pipeline of TTT-NTP.} At each adapted MLP block, the
  current gated activation $\bm{z}_{\ell,t}$ is the key. The write target is
  the next position's same-layer contextual state $\bm{h}_{\ell,t+1}$, passed
  through a small learned linear projection $\bm{W}_{\ell}^{\mathrm{proj}}$ before being written into
  the down-projection fast weight. Writes are accumulated causally as an
  exclusive chunk prefix sum.}
  \label{fig:pipeline}
\end{figure*}

Following~\citet{feng2026place}, we instantiate TTT-NTP in selected SwiGLU MLP blocks inside the transformer layers (\cref{fig:pipeline}). Let
$\bm{h}_{\ell,t}\in\mathbb{R}^{d}$ be the post-normalization input to the MLP
block at layer $\ell$ and position $t$. The pretrained MLP computes:
\begin{align}
\bm{z}_{\ell,t}
&=
\phi\!\left(\bm{W}_{\ell}^{\mathrm{gate}}\bm{h}_{\ell,t}\right)
\odot
\left(\bm{W}_{\ell}^{\mathrm{up}}\bm{h}_{\ell,t}\right)
\in \mathbb{R}^{d_{\mathrm{ff}}},
\label{eq:mlp-z}
\\
\bm{o}_{\ell,t}
&=
\bm{W}_{\ell}^{\mathrm{down}}\bm{z}_{\ell,t}
\in \mathbb{R}^{d},
\label{eq:mlp-o}
\end{align}
where $\phi$ is SiLU and $\bm{W}_{\ell}^{\mathrm{down}}\in
\mathbb{R}^{d\times d_{\mathrm{ff}}}$. We place one fast weight at each adapted layer $\ell\in\mathcal{A}$ by reusing the down-projection itself, i.e., $\bm{W}_{\ell}=\bm{W}_{\ell}^{\mathrm{down}}$. The key vector is the current gated activation $\bm{z}_{\ell,t}$, and
the write to the fast weight at position $t$ is the negative-gradient step on the inner-loop loss of \cref{eq:ttt-update}, which has a rank-one form: %\jh{There is a gap here. What is $\Delta \bm{W}_{\ell,t}$? How do you derive it? Is it from $\Delta \bm{W}_{\ell,t} = \nabla_{\bm{W}}\,  \mathcal{L}_2\!\bigl(\bm{W}\bm{k}_{t},\,\bm{v}_{t}\bigr)$}
\begin{equation}
\Delta \bm{W}_{\ell,t}
:= -\nabla_{\bm{W}_{\ell}}\,  \mathcal{L}\!\bigl(\bm{W}_{\ell}\bm{z}_{\ell,t},\,\bm{v}_{\ell,t}\bigr)=
\bm{v}_{\ell,t}\bm{z}_{\ell,t}^{\top}.
\label{eq:single-write}
\end{equation}

The TTT-NTP value target is derived from the same layer's next contextual hidden state, $\bm{h}_{\ell,t+1}$. This state is computed under the causal mask in the same forward pass after the model observes token $x_{t+1}$, and serves as the value target for the fast-weight write. We keep a lightweight learned linear projection
$\bm{W}_{\ell}^{\mathrm{proj}}\in\mathbb{R}^{d\times d}$ between the target state and the
down-projection fast weight: %\jh{During TTT, what are the learnable weights? $\bm{W}_{\ell}^{\mathrm{proj}}$ or $\bm{W}_{\ell}^{\mathrm{down}}$ or both? Which one should we call as the fast weight? I thought it would be $\bm{W}_{\ell}^{\mathrm{down}}$. Are you also updating $\bm{W}_{\ell}^{\mathrm{proj}}$ by the gradient of the inner TTT NTP loss w.r.t. $\bm{W}_{\ell}^{\mathrm{proj}}$?}
\begin{equation}
\bm{v}_{\ell,t}^{\mathrm{NTP}}
=
\bm{W}_{\ell}^{\mathrm{proj}}\bm{h}_{\ell,t+1}.
\label{eq:ntp-target}
\end{equation}
By instantiating $\bm{v}_{\ell,t}=\bm{v}_{\ell,t}^{\mathrm{NTP}}$ in \cref{eq:single-write}, the TTT-NTP single-position write to $\bm{W}_{\ell}$ is
\begin{equation}
\Delta \bm{W}_{\ell,t}^{\mathrm{NTP}}
:=
\bigl(\bm{W}_{\ell}^{\mathrm{proj}}\bm{h}_{\ell,t+1}\bigr)\bm{z}_{\ell,t}^{\top}.
\label{eq:single-ntp-write}
\end{equation}
We initialize $\bm{W}_{\ell}^{\mathrm{proj}}$ and train it jointly
with the backbone during continual pretraining. The outer loss remains the standard
next-token cross-entropy; the inner-loop target in \cref{eq:ntp-target} is
the per-layer next-token state transition induced by that next-token objective. %\jh{Why do you call it ``layer-local''? Is there a better way to call it?} 
After continual pretraining, $\bm{W}_{\ell}^{\mathrm{proj}}$ is frozen for the test-time training phase.

This choice removes the need to aggregate the target over a neighborhood of
positions. Prior recipes do exactly that, typically through a learned
convolution that combines hidden states at several positions around $t$
into a value vector. In TTT-NTP, context enters through the causal forward
pass itself: $\bm{h}_{\ell,t+1}$ is already the model's contextual
representation after the next observed token, and
$\bm{W}_{\ell}^{\mathrm{proj}}$ is only a pointwise linear projection of that
single state into the fast-weight write space. The resulting write pushes
the layer's MLP output along the same trajectory the model itself follows
under next-token prediction, so the fast weight stores predictive state
rather than a learned reconstruction proxy.

\subsection{Chunk-Parallel Causal Update}
\label{sec:method:chunk}

To preserve throughput, writes are accumulated at the chunk level. We partition the long sequence into chunks of length $K$, and define a list of the token positions in a chunk $c$ as:
\begin{equation}
\mathcal{T}_{c}
=
[(c-1)K+1,\ldots,cK],
~~ \forall c\in [1, C].
\label{eq:chunk}
\end{equation}
Because position $t$ is paired with the next position $t+1$, we only consider
within-chunk token pairs for fast weight update; thus, we exclude the last token from the updates.
\begin{equation}
\mathcal{T}_{c}^{-}
=[(c-1)K+1,\ldots,cK-1],~~\forall c\in[1, C].
\label{eq:chunk-minus}
\end{equation}
% \jh{Where do you use the two sets for? These two sets are defined here but not used elsewhere.}
The per-chunk NTP accumulator to the fast weight is computed as:
\begin{equation}
\Delta \bm{W}_{\ell}^{(c)}
:=
\sum_{t\in\mathcal{T}_{c}^{-}} \Delta \bm{W}_{\ell,t}^{\mathrm{NTP}}
=\bm{W}_{\ell}^{\mathrm{proj}}
\sum_{t\in\mathcal{T}_{c}^{-}}
\bm{h}_{\ell,t+1}\bm{z}_{\ell,t}^{\top}
\in \mathbb{R}^{d\times d_{\mathrm{ff}}}.
\label{eq:chunk-write}
\end{equation}
Since TTT-NTP updates the fast weight $\bm{W}_{\ell}$ chunk by chunk, it writes per-chunk accumulators to $\bm{W}_{\ell}$ sequentially. Let $\bm{W}_{\ell}^{(c)}$ denote the snapshot of the fast weight used to process chunk $c$. $\bm{W}_{\ell}^{(1)}$ is reusing $\bm{W}_{\ell}^{\mathrm{down}}$ in place at chunk 1, and then $\bm{W}_{\ell}^{(c)}$ is computed as the exclusive prefix sum of updates from all preceding chunks:
\begin{align}
\bm{W}_{\ell}^{(1)}
&=
\bm{W}_{\ell}^{\mathrm{down}}, \\
\bm{W}_{\ell}^{(c)}
&=
\bm{W}_{\ell}^{\mathrm{down}}
+
\eta
\sum_{c'<c}
\Delta \bm{W}_{\ell}^{(c')}.
\label{eq:prefix-weight}
\end{align}
Note that each position $t\in\mathcal{T}_{c}$ then uses
$\bm{W}_{\ell}^{(c)}\bm{z}_{\ell,t}$ in place of the original
$\bm{W}_{\ell}^{\mathrm{down}}\bm{z}_{\ell,t}$ in \cref{eq:mlp-o}.

The update is causal. Across chunks, chunk $c$ can only read writes from
chunks $c'<c$. Within a chunk, all positions share the same fast weight,
which excludes writes produced by that chunk. The restricted set
$\mathcal{T}_{c}^{-}$ prevents a write from crossing into the next chunk.
Although the target for position $t$ is $\bm{h}_{\ell,t+1}$, the resulting
write never affects the computation of position $t+1$ itself; it is available
only for later chunks or later inference-time generation. Thus the method
uses observed next tokens as test-time supervision without leaking future
information into their own predictions.

\subsection{Inference-Time Closed-Form Write}
\label{sec:method:cf}

The chunk-parallel update in \cref{eq:chunk-write} sums bare outer products
and drops the residual term that would appear in a literal least-squares
regression step. For a single position $t$, the gradient of
the regression loss $\frac{1}{2}\|\bm{W}\bm{z}_{\ell,t}-\bm{v}_{\ell,t}^{\mathrm{NTP}}\|^{2}$ w.r.t. $\bm{W}$ is
\begin{equation}
\nabla_{\bm{W}}\,
    \mathcal{L}_2\!\bigl(\bm{W}\bm{z}_{\ell,t},\,\bm{v}_{\ell,t}^{\mathrm{NTP}}\bigr) = \bm{W}\bm{z}_{\ell,t}\bm{z}_{\ell,t}^{\top}
-\bm{v}_{\ell,t}^{\mathrm{NTP}}\bm{z}_{\ell,t}^{\top},
\label{eq:gd-residual}
\end{equation}
An exact regression step would descend this gradient, i.e.\ add
$\bm{v}_{\ell,t}^{\mathrm{NTP}}\bm{z}_{\ell,t}^{\top}-\bm{W}\bm{z}_{\ell,t}\bm{z}_{\ell,t}^{\top}$;
the chunk write keeps only the target outer product
$\bm{v}_{\ell,t}^{\mathrm{NTP}}\bm{z}_{\ell,t}^{\top}$ and drops the
$\bm{W}$-dependent term, which is not prefix-summable. This is exactly the
simplification that makes the chunk update parallel across positions, and it
recovers the inner-product write of \cref{eq:single-write}.

Training and inference therefore optimize the same next-position target with
\emph{different} objectives, for a deliberate reason: the per-token training
write must drop the $\bm{W}$-dependent term to stay prefix-summable, but the
prompt is observed in full at inference, so we can afford to solve the complete
squared-error regression (\cref{eq:cf-objective}) once and restore that term.
\Cref{sec:experiments:cf-ablation} confirms this design choice.

Given a prompt $x_{1:T}$, we run one forward pass and cache, at each adapted
layer $\ell$, the gated MLP intermediates $\{\bm{z}_{\ell,t}\}_{t=1}^{T}$ and
the MLP-input hidden states $\{\bm{h}_{\ell,t}\}_{t=1}^{T}$. In practice, for
efficiency on long prompts, we fit the regression on a suffix of at most
$8{,}192$ prefill tokens; the formulation is otherwise the same as if we
used the full prompt. Using the same next-position alignment as in training,
we stack
\begin{align}
\bm{X}_{\ell}
&=
[\bm{z}_{\ell,1},\dots,\bm{z}_{\ell,T-1}]
\in \mathbb{R}^{d_{\mathrm{ff}}\times(T-1)},
\label{eq:cf-X}
\\
\bm{Y}_{\ell}
&=
\bm{W}_{\ell}^{\mathrm{proj}}
[\bm{h}_{\ell,2},\dots,\bm{h}_{\ell,T}]
\in \mathbb{R}^{d\times(T-1)}.
\label{eq:cf-Y}
\end{align}
Each column of $\bm{Y}_{\ell}$ is the same per-position value target
$\bm{v}_{\ell,t}^{\mathrm{NTP}}=\bm{W}_{\ell}^{\mathrm{proj}}\bm{h}_{\ell,t+1}$ used
during pretraining. Training and inference therefore share a single target
form; the closed-form solve fits one $\Delta \bm{W}$ to that target jointly
over the prompt, instead of accumulating per-position outer products.

We fit a prompt-specific perturbation of the down-projection by ridge
regression:
\begin{equation}
\min_{\Delta \bm{W}}\;
\bigl\|\bm{Y}_{\ell} - (\bm{W}_{\ell}^{\mathrm{down}}+\Delta\bm{W})\bm{X}_{\ell}\bigr\|_{F}^{2}
+ \lambda\,\|\Delta \bm{W}\|_{F}^{2},
\label{eq:cf-objective}
\end{equation}
with closed form
\begin{equation}
\Delta \bm{W}_{\ell}^{\mathrm{CF}}
=
(\bm{Y}_{\ell}-\bm{W}_{\ell}^{\mathrm{down}}\bm{X}_{\ell})
\bm{X}_{\ell}^{\top}
(\bm{X}_{\ell}\bm{X}_{\ell}^{\top}+\lambda\bm{I})^{-1},
\label{eq:cf-write}
\end{equation}
or its dual form when $T<d_{\mathrm{ff}}$. This regression is meaningful only because the down-projection was shaped during TTT-NTP pretraining to be read in this way.

We apply $\bm{W}_{\ell}^{\mathrm{down}}\leftarrow\bm{W}_{\ell}^{\mathrm{down}}+\eta\,\Delta\bm{W}_{\ell}^{\mathrm{CF}}$, decode the answer, and restore the original weight after each sample. Because we reuse the original prompt key--value cache, the write affects decode-time tokens without recomputing prompt activations under the modified MLP.

\section{Experiments}
\label{sec:experiments}

We evaluate Test-Time Training with Next-Token Prediction (TTT-NTP) from three angles. We first measure long-context retrieval, where the next-position signal should matter most, on the synthetic RULER suite, and then ask whether the same gains carry over to real-world long-document question answering on LongBench-v2. We next isolate \emph{what} drives the gains through controlled ablations of the supervisory target and the inference-time write. Finally, we verify on standard commonsense and knowledge benchmarks that the long-context improvements do not come at a cost to the base model's general capability.

\subsection{Experimental Setup}
\label{sec:experiments:setup}

\paragraph{Backbones.} We evaluate TTT-NTP across four open backbones from three model families spanning a $0.6$--$8$B size range, so the results probe generality rather than a single model: Llama-3.1-8B-Base \citep{grattafiori2024llama}, Mistral-7B-v0.3 \citep{jiang2023mistral}, Qwen3-4B-Base, and Qwen3-0.6B-Base \citep{yang2025qwen3}. All are full-attention decoders with SwiGLU MLPs.

\paragraph{Continual pretraining (CPT).} All trained variants share an identical recipe: continual pretraining on long-document text from the pretraining split of Long-Data-Collections at $32{,}768$-token sequence length, with a per-backbone token budget of $0.4$B tokens for Llama-3.1-8B, $0.1$B for Mistral-7B-v0.3, $2$B for Qwen3-4B, and $0.2$B for Qwen3-0.6B. The CPT baseline applies this recipe without fast-weight updates; In-Place TTT and TTT-NTP share the same data, optimizer, compute, fast-weight placement, chunk size, and inner-loop learning rate, so the only difference between them is the fast-weight target. Full hyperparameters appear in \cref{sec:appendix:training}.

\paragraph{Evaluation.} Long-context retrieval is measured by RULER \citep{hsieh2024ruler} Full-13, the official $13$-task aggregate over needle-in-a-haystack retrieval variants, variable tracking, frequent- and common-word extraction, and question answering, at context lengths $4$k, $8$k, $16$k, and $32$k. All rows use the same RULER evaluation harness; each adaptive method applies its specified inference-time update. We further assess real-world long-document QA on the medium split of LongBench-v2 \citep{bai2025longbenchv2deeperunderstanding}, $215$ multiple-choice questions over documents of $33$k to $128$k words across six task domains, evaluated under a common $32$k-token context budget with head-and-tail truncation. For general capability we evaluate the trained backbone on HellaSwag \citep{zellers2019hellaswag}, ARC \citep{clark2018think}, PIQA \citep{bisk2020piqa}, and MMLU \citep{hendrycks2020measuring}, under the standard lm-evaluation-harness protocols.

\subsection{Baselines}
\label{sec:experiments:baselines}

We compare against four points spanning learned-update and extra inference-time adaptation settings:
\begin{itemize}
  \item \textbf{Base}: the released backbone used directly, without any continual pretraining or inference-time adaptation, under the same RULER evaluation harness.
  \item \textbf{query-side TTT (qTTT)} \citep{bansal2025letsnotjustthings}: an inference-only baseline that, before answering, takes a few self-supervised gradient steps on sampled spans of the prompt while reusing the cached keys and values, applied on top of Base with $32$ inner steps on $128$-token spans, learning rate $1\times10^{-5}$, weight decay $0.01$, and gradient clipping $1.0$.
  \item \textbf{CPT}: continual pretraining with no fast-weight TTT, following the recipe of \cref{sec:experiments:setup}. Our primary no-TTT baseline.
  \item \textbf{In-Place TTT} \citep{feng2026place}: continual pretraining with the published fast-weight TTT recipe, whose inner-loop target is a small learned convolution over a local window of hidden states around each position, matched to our method in budget, optimizer, fast-weight placement, chunk size.
\end{itemize}
Our method is TTT-NTP, which combines the chunk-parallel fast-weight update of \cref{sec:method:ntp,sec:method:chunk} with the closed-form inference-time write of \cref{sec:method:cf}. \Cref{tab:ruler-main} reports its long-context retrieval scores.

\subsection{Long-Context Retrieval Results}
\label{sec:experiments:ruler}

\Cref{tab:ruler-main} compares TTT-NTP with all baselines on RULER Full-13 across the four backbones.

\begin{table*}[t]
\centering
\small
\renewcommand{\arraystretch}{1.15}
\begin{tabularx}{\textwidth}{l *{6}{>{\centering\arraybackslash}X}}
\toprule
Model & Tokens & 4k & 8k & 16k & 32k & Avg \\
\midrule
\multicolumn{7}{l}{\textit{Llama-3.1-8B-Base}} \\
\midrule
Base         & 0    & \textbf{65.05} & 56.98 & 46.94 & 54.22 & 55.80 \\
qTTT         & 0    & 64.72\dd{0.33} & 56.78\dd{0.20} & 47.55\du{0.61} & 55.10\du{0.88} & 56.04\du{0.24} \\
CPT          & 0.4B & 55.09\dd{9.96} & 51.14\dd{5.84} & 46.61\dd{0.33} & 46.93\dd{7.29} & 49.94\dd{5.86} \\
In-Place TTT & 0.4B & 57.82\dd{7.23} & 52.65\dd{4.33} & 47.23\du{0.29} & 49.34\dd{4.88} & 51.76\dd{4.04} \\
\rowcolor{ourcolor}\textbf{TTT-NTP} & 0.4B & 63.39\dd{1.66} & \textbf{59.77}\du{2.79} & \textbf{53.90}\du{6.96} & \textbf{61.73}\du{7.51} & \textbf{59.70}\du{3.90} \\
\midrule
\multicolumn{7}{l}{\textit{Mistral-7B-v0.3}} \\
\midrule
Base         & 0     & 73.26 & 76.76 & 72.24 & 60.13 & 70.60 \\
qTTT         & 0     & 68.59\dd{4.67} & 70.63\dd{6.13} & 67.09\dd{5.15} & 57.23\dd{2.90} & 65.89\dd{4.71} \\
CPT          & 0.1B & 69.25\dd{4.01} & 68.44\dd{8.32} & 61.88\dd{10.36} & 52.15\dd{7.98} & 62.93\dd{7.67} \\
In-Place TTT & 0.1B & 68.41\dd{4.85} & 66.61\dd{10.15} & 62.27\dd{9.97} & 51.31\dd{8.82} & 62.15\dd{8.45} \\
\rowcolor{ourcolor}\textbf{TTT-NTP} & 0.1B & \textbf{81.51}\du{8.25} & \textbf{79.13}\du{2.37} & \textbf{72.64}\du{0.40} & \textbf{61.23}\du{1.10} & \textbf{73.63}\du{3.03} \\
\midrule
\multicolumn{7}{l}{\textit{Qwen3-4B-Base}} \\
\midrule
Base         & 0  & 93.23 & 84.03 & 74.96 & 75.47 & 81.92 \\
qTTT         & 0  & \textbf{93.75}\du{0.52} & 84.88\du{0.85} & 75.34\du{0.38} & 76.18\du{0.71} & 82.53\du{0.61} \\
CPT          & 2B & 88.27\dd{4.96} & 82.93\dd{1.10} & 71.32\dd{3.64} & 68.31\dd{7.16} & 77.71\dd{4.21} \\
In-Place TTT & 2B & 87.98\dd{5.25} & 84.71\du{0.68} & 70.04\dd{4.92} & 66.59\dd{8.88} & 77.33\dd{4.59} \\
\rowcolor{ourcolor}\textbf{TTT-NTP} & 2B & 89.69\dd{3.54} & \textbf{87.43}\du{3.40} & \textbf{86.70}\du{11.74} & \textbf{80.08}\du{4.61} & \textbf{85.98}\du{4.06} \\
\midrule
\multicolumn{7}{l}{\textit{Qwen3-0.6B-Base}} \\
\midrule
Base         & 0     & 81.83 & 70.95 & 65.24 & 56.20 & 68.55 \\
qTTT         & 0     & 81.99\du{0.16} & 71.57\du{0.62} & 65.43\du{0.19} & 56.74\du{0.54} & 68.93\du{0.38} \\
CPT          & 0.2B & 82.08\du{0.25} & 70.33\dd{0.62} & 65.63\du{0.39} & 57.38\du{1.18} & 68.86\du{0.31} \\
In-Place TTT & 0.2B & 72.41\dd{9.42} & 68.40\dd{2.55} & 67.57\du{2.33} & \textbf{59.18}\du{2.98} & 66.89\dd{1.66} \\
\rowcolor{ourcolor}\textbf{TTT-NTP} & 0.2B & \textbf{82.71}\du{0.88} & \textbf{73.31}\du{2.36} & \textbf{71.39}\du{6.15} & 58.30\du{2.10} & \textbf{71.43}\du{2.88} \\
\bottomrule
\end{tabularx}
\caption{RULER Full-13 accuracy across four backbones (Llama-3.1-8B-Base, Mistral-7B-v0.3, Qwen3-4B-Base, and Qwen3-0.6B-Base; blocks ordered by model size). Subscript arrows give the absolute percentage-point change relative to Base for each backbone ({\textcolor{deltaup}{$\uparrow$}}~improvement, {\textcolor{deltadown}{$\downarrow$}}~decline); the best result in each column is in \textbf{bold}. The \emph{Tokens} column gives the continual-pretraining budget. All rows use the same RULER evaluation harness; In-Place TTT and TTT-NTP share the same fast-weight placement, chunk size, and inner-loop learning rate.}
\label{tab:ruler-main}
\end{table*}

\textbf{Consistent gains across backbones.} TTT-NTP is the only adaptation that improves the released model on every backbone, whereas the alternatives are unreliable: continued pretraining and the convolutional In-Place target frequently \emph{degrade} long-context retrieval, and inference-time qTTT helps little or hurts. Since In-Place TTT is trained on the same data and compute and differs only in its supervisory target, the gap between the two attributes the improvement to next-position supervision itself rather than to additional tokens or the rank-one write.

\textbf{Largest where retrieval is hardest.} On three of the four backbones the improvement grows with context length and peaks at $16$k to $32$k, reaching $+11.7$ at $16$k on Qwen3-4B, at the cost of a small dip at $4$k on Llama-3.1-8B and Qwen3-4B that trades short-range for long-range capacity; Mistral-7B-v0.3 instead gains most at $4$k.

\textbf{Robust to backbone strength and scale.} On a weak long-context base, where naive adaptation instead backfires, TTT-NTP adds nearly four points; on the strongest bases---where CPT and In-Place TTT shed several points and qTTT either regresses or barely moves the average---it is the only method that still improves. The same ranking holds from $8$B down to $0.6$B.

\FloatBarrier
\subsection{Real-World Long-Document QA}
\label{sec:experiments:lbv2}

RULER is synthetic. To test whether the same inference-time write transfers to naturally occurring long documents, we additionally evaluate on LongBench-v2 \citep{bai2025longbenchv2deeperunderstanding}, a multiple-choice long-document QA benchmark spanning six task domains. We report its \emph{medium}-length split of $215$ questions over documents of $33$k to $128$k words under a common $32$k-token context budget; inputs longer than the budget keep their head and tail, so every backbone is compared at the same effective input length. \Cref{tab:lbv2} breaks accuracy down by domain.

\textbf{Transfer to real long documents.} TTT-NTP attains the best overall accuracy on both backbones, $+5.6$ over Base on Llama-3.1-8B and $+3.7$ on Mistral-7B-v0.3, ahead of CPT, In-Place TTT, and qTTT; the next-position write thus helps on naturally occurring long-document QA, not only on synthetic needle retrieval.

\textbf{Retrieval-oriented domains drive the gain.} Single- and multi-document QA and long structured-data understanding account for most of the improvement, with Mistral rising from $13.0$ to $30.4$ on structured data, mirroring the RULER results, whereas the smallest splits, code with $9$ questions and dialogue with $19$, are too small to read much into.

\begin{table*}[t]
  \centering
  \small
  \setlength{\tabcolsep}{3pt}
  \begin{tabularx}{\textwidth}{l *{7}{>{\centering\arraybackslash}X}}
    \toprule
    Method & Single-Doc & Multi-Doc & ICL & Dialogue & Code & Struct & Overall \\
    \midrule
    \multicolumn{8}{l}{\textit{Llama-3.1-8B-Base}} \\
    \midrule
    Base         & 26.0 & 25.0 & 27.9 & 15.8 & 33.3 & 26.1 & 25.6 \\
    CPT          & 31.2\du{5.2} & 27.3\du{2.3} & 25.6\dd{2.3} & 31.6\du{15.8} & 44.4\du{11.1} & 26.1 & 29.3\du{3.7} \\
    In-Place TTT & 35.1\du{9.1} & 20.5\dd{4.5} & 27.9 & 21.1\du{5.3} & 44.4\du{11.1} & 26.1 & 28.8\du{3.2} \\
    qTTT         & 28.6\du{2.6} & 25.0 & 27.9 & 15.8 & 33.3 & 26.1 & 26.5\du{0.9} \\
    \rowcolor{ourcolor}\textbf{TTT-NTP (Ours)} & \textbf{33.8}\du{7.8} & \textbf{29.5}\du{4.5} & \textbf{30.2}\du{2.3} & \textbf{21.1}\du{5.3} & \textbf{33.3} & \textbf{34.8}\du{8.7} & \textbf{31.2}\du{5.6} \\
    \midrule
    \multicolumn{8}{l}{\textit{Mistral-7B-v0.3}} \\
    \midrule
    Base         & 29.9 & 25.0 & 25.6 & 31.6 & 33.3 & 13.0 & 26.5 \\
    CPT          & 35.1\du{5.2} & 25.0 & 25.6 & 10.5\dd{21.1} & 33.3 & 21.7\du{8.7} & 27.4\du{0.9} \\
    In-Place TTT & 32.5\du{2.6} & 29.5\du{4.5} & 25.6 & 15.8\dd{15.8} & 33.3 & 17.4\du{4.4} & 27.4\du{0.9} \\
    qTTT         & 31.2\du{1.3} & 25.0 & 18.6\dd{7.0} & 31.6 & 22.2\dd{11.1} & 13.0 & 25.1\dd{1.4} \\
    \rowcolor{ourcolor}\textbf{TTT-NTP (Ours)} & \textbf{32.5}\du{2.6} & \textbf{31.8}\du{6.8} & \textbf{23.3}\dd{2.3} & \textbf{31.6} & \textbf{33.3} & \textbf{30.4}\du{17.4} & \textbf{30.2}\du{3.7} \\
    \bottomrule
  \end{tabularx}
  \caption{\textbf{LongBench-v2 medium split, by task domain.} Multiple-choice accuracy on the $215$ medium questions ($33$k--$128$k words), evaluated under a common $32$k-token context budget (head$+$tail truncation). Per-domain question counts: Single-Doc QA $77$, Multi-Doc QA $44$, In-context Learning (ICL) $43$, Dialogue History $19$, Code Repository $9$, Structured Data $23$. Subscripts give the per-domain change relative to Base for each backbone ({\textcolor{deltaup}{$\uparrow$}}~improvement, {\textcolor{deltadown}{$\downarrow$}}~decline); TTT-NTP (Ours) values are in \textbf{bold}.}
  \label{tab:lbv2}
\end{table*}

\FloatBarrier
\subsection{Ablation Study}
\label{sec:experiments:ablation}

\subsubsection{Choice of Supervision Target}
\label{sec:experiments:target-ablation}

TTT-NTP supervises the inner-loop write with the next position's same-layer contextual hidden state $\bm{h}_{\ell,t+1}$. To isolate the contribution of this specific choice---and not the rank-one mechanism, the placement, or the chunk-parallel schedule---we ablate the target while holding the training data, token budget, optimizer, fast-weight placement, chunk size, inner-loop learning rate, and update mechanism fixed at the values of \cref{sec:experiments:setup}, varying only how the layer-local target is constructed:
\begin{itemize}
  \item \textbf{Past-5}: target aggregated from the five preceding positions $\{\bm{h}_{\ell,t-1},\ldots,\bm{h}_{\ell,t-5}\}$ via a learned causal convolution.
  \item \textbf{Next-5}: target aggregated from the five following positions $\{\bm{h}_{\ell,t+1},\ldots,\bm{h}_{\ell,t+5}\}$ via a learned forward convolution.
  \item \textbf{Bi-dir-5}: target aggregated from the symmetric $11$-position window $\{\bm{h}_{\ell,t-5},\ldots,\bm{h}_{\ell,t+5}\}$ via a learned bidirectional convolution. This is a symmetric convolutional value-proxy baseline inspired by local-target TTT recipes \citep{feng2026place}.
  \item \textbf{TTT-NTP (Ours)}: target is the single next position's hidden state $\bm{h}_{\ell,t+1}$, with no convolution; the signal enters the rank-one write through the learned linear projection $\bm{W}_{\ell}^{\mathrm{proj}}$ (\cref{eq:ntp-target}).
\end{itemize}

\begin{figure}[t]
  \centering
  \includegraphics[width=\linewidth]{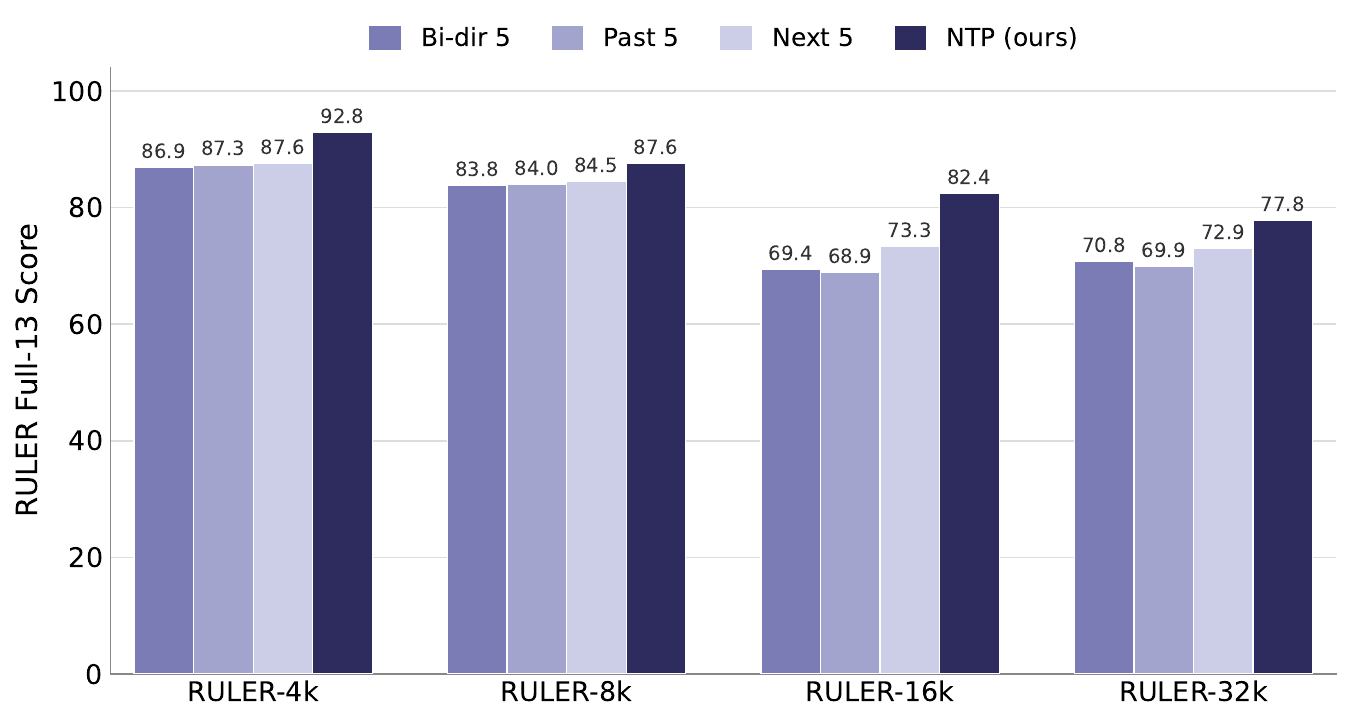}
  \caption{\textbf{Target ablation on RULER Full-13 (Qwen3-4B-Base).} All four variants share the same training data, token budget, fast-weight placement, chunk size, inner-loop learning rate, and rank-one update mechanism; only the layer-local target differs. \emph{Past-5} and \emph{Next-5} aggregate five preceding or following positions through a learned unidirectional convolution; \emph{Bi-dir-5} aggregates a symmetric $11$-position window through a bidirectional convolution. \emph{TTT-NTP (ours)} writes the single next position $\bm{h}_{\ell,t+1}$.}
  \label{fig:target-ablation}
\end{figure}

\Cref{fig:target-ablation} shows that the single next-position target outperforms every convolutional aggregation at all four evaluated lengths. The gap is already at least five points at $4$k, where the convolutional targets reach $86.9$ to $87.6$ against $92.8$ for TTT-NTP, grows to roughly nine points at $16$k, $68.9$ to $73.3$ against $82.4$, and remains about five points at $32$k, $69.9$ to $72.9$ against $77.8$. The three convolutional aggregations cluster together at every length: smoothing the target through a learned convolution, in any direction, hurts long-context retrieval relative to writing the next contextual state directly. This ablation pins the gain of \cref{tab:ruler-main} to the supervisory target rather than the rank-one mechanism or the local context window.

\subsubsection{Closed-Form Inference Write: Solver and Regularization}
\label{sec:experiments:cf-ablation}

The inference-time write of \cref{eq:cf-write} is the ridge least-squares solution for the prompt-specific perturbation: over the fit window it minimizes $\|\bm{Y}_{\ell}-(\bm{W}_{\ell}^{\mathrm{down}}+\Delta\bm{W})\bm{X}_{\ell}\|_F^2+\lambda\|\Delta\bm{W}\|_F^2$, where $\bm{X}_{\ell}$ stacks the cached keys and $\bm{Y}_{\ell}$ the next-position targets. To isolate the role of the \emph{solver}---not the trained target, placement, or perturbation scale $\eta$, all held fixed---we compare three ways of mapping the residual $\bm{R}_{\ell}=\bm{Y}_{\ell}-\bm{W}_{\ell}^{\mathrm{down}}\bm{X}_{\ell}$ to a write.

Writing this closed form in terms of the residual makes the solver explicit:
\begin{equation}
\Delta\bm{W}^{\mathrm{CF}}_{\ell}
= \underbrace{\bm{R}_{\ell}\bm{X}_{\ell}^{\top}}_{\textstyle\text{residual--key correlation}}\,
\underbrace{\bigl(\bm{X}_{\ell}\bm{X}_{\ell}^{\top}+\lambda\bm{I}\bigr)^{-1}}_{\textstyle\text{key whitening}}.
\label{eq:inner-vs-ridge}
\end{equation}
The numerator $\bm{R}_{\ell}\bm{X}_{\ell}^{\top}$ is a Hebbian, correlational write---the same outer-product form as the inner-product (linear) training loss in \cref{eq:ttt-update}---and the Gram inverse whitens it by the key second moment. The two ablations each switch off one of these factors: \emph{Inner-product} keeps only the un-whitened correlation $\bm{R}_{\ell}\bm{X}_{\ell}^{\top}$, and \emph{no regularization} keeps the inverse but sets $\lambda\!\to\!0$. The inner-product write is exactly the per-token training-time update of \cref{eq:ttt-update} applied once over the whole prompt; testing it at inference therefore asks whether the training loss can be reused for the one-shot write. \Cref{tab:cf-ablation} reports all three against each backbone's released Base.

\textbf{Whitening is decisive.} Dropping the Gram inverse collapses retrieval at every length, with the average falling to $71.22$ on Qwen3-4B and $12.61$ on Llama-3.1-8B---far below Base. The factor $(\bm{X}_{\ell}\bm{X}_{\ell}^{\top}+\lambda\bm{I})^{-1}$ rescales the write to equalize the key directions; without it the Hebbian write is amplified along the highest-variance (most frequent or repeated) keys rather than solving for the target, so a few directions dominate the patched down-projection and the outputs degenerate. This is the same Hebbian form TTT-NTP trains with: dense, small per-position writes are stable during co-adapted chunk-parallel training, but the identical un-whitened form fails as a single one-shot prompt write---which is exactly why inference solves the full squared-error objective rather than reusing the training loss.

\textbf{Regularization matters.} Keeping the whitening while setting $\lambda\!\to\!0$ stays close to the full solve and well above the inner-product write, yet only the complete regularized ridge (Ours) improves over Base at every length beyond $4$k on both backbones. Both the whitening and a non-zero $\lambda$ are therefore needed for a stable one-shot prompt write.

\begin{table}[t]
  \centering
  \small
  \setlength{\tabcolsep}{3pt}
  \begin{tabularx}{\linewidth}{@{}l *{5}{>{\centering\arraybackslash}X}@{}}
    \toprule
    Inference write & 4k & 8k & 16k & 32k & Avg \\
    \midrule
    \multicolumn{6}{@{}l}{\textit{Llama-3.1-8B}} \\
    \midrule
    Base & 65.05 & 56.98 & 46.94 & 54.22 & 55.80 \\
    Inner-product & 7.19\dd{57.86} & 9.86\dd{47.12} & 20.76\dd{26.18} & 12.63\dd{41.59} & 12.61\dd{43.19} \\
    Ridge, no reg. & 62.81\dd{2.24} & 56.76\dd{0.22} & 51.74\du{4.80} & 53.13\dd{1.09} & 56.11\du{0.31} \\
    \rowcolor{ourcolor}\textbf{Ridge (Ours)} & \textbf{63.39}\dd{1.66} & \textbf{59.77}\du{2.79} & \textbf{53.90}\du{6.96} & \textbf{61.73}\du{7.51} & \textbf{59.70}\du{3.90} \\
    \midrule
    \multicolumn{6}{@{}l}{\textit{Qwen3-4B}} \\
    \midrule
    Base & 93.23 & 84.03 & 74.96 & 75.47 & 81.92 \\
    Inner-product & 75.96\dd{17.27} & 75.79\dd{8.24} & 67.75\dd{7.21} & 65.38\dd{10.09} & 71.22\dd{10.70} \\
    Ridge, no reg. & 89.52\dd{3.71} & \textbf{87.89}\du{3.86} & 86.29\du{11.33} & 79.98\du{4.51} & 85.92\du{4.00} \\
    \rowcolor{ourcolor}\textbf{Ridge (Ours)} & \textbf{89.69}\dd{3.54} & 87.43\du{3.40} & \textbf{86.70}\du{11.74} & \textbf{80.08}\du{4.61} & \textbf{85.98}\du{4.06} \\
    \bottomrule
  \end{tabularx}
  \caption{\textbf{Closed-form inference-write ablation (RULER Full-13, per length).} Varying only the solver for the prompt write $\Delta\bm{W}$; subscripts are the change vs.\ Base ({\textcolor{deltaup}{$\uparrow$}}/{\textcolor{deltadown}{$\downarrow$}}), best write rule per column in \textbf{bold}. The Hebbian inner-product write (no Gram whitening) collapses, and dropping the ridge regularizer also hurts; only the full regularized solve (Ours) improves over Base.}
  \label{tab:cf-ablation}
\end{table}

\FloatBarrier
\subsection{General Capability Evaluation}
\label{sec:experiments:general}

\Cref{tab:general} evaluates each TTT-NTP-trained backbone on standard commonsense and knowledge benchmarks, to check whether the long-context retrieval gains of \cref{tab:ruler-main} come at a cost to general capability.

\begin{table}[!ht]
\centering
\small
\setlength{\tabcolsep}{3pt}
\begin{tabular}{ll cccccc}
\toprule
Backbone & Method & HellaSwag & ARC-e & ARC-c & PIQA & MMLU & Avg \\
\midrule
\multirow{2}{*}{Llama-3.1-8B} & Base    & 71.15 & 80.40 & 58.02 & 83.40 & 63.45 & 71.28 \\
 & \cellcolor{ourcolor}\textbf{TTT-NTP} & \cellcolor{ourcolor}\textbf{71.50}\du{0.35} & \cellcolor{ourcolor}\textbf{80.80}\du{0.40} & \cellcolor{ourcolor}\textbf{58.70}\du{0.68} & \cellcolor{ourcolor}83.40 & \cellcolor{ourcolor}63.37\dd{0.08} & \cellcolor{ourcolor}\textbf{71.56}\du{0.27} \\
\midrule
\multirow{2}{*}{Mistral-7B-v0.3} & Base    & 72.85 & 78.20 & 60.24 & 84.20 & 59.64 & 71.03 \\
 & \cellcolor{ourcolor}\textbf{TTT-NTP} & \cellcolor{ourcolor}69.05\dd{3.80} & \cellcolor{ourcolor}\textbf{78.40}\du{0.20} & \cellcolor{ourcolor}59.64\dd{0.60} & \cellcolor{ourcolor}\textbf{85.00}\du{0.80} & \cellcolor{ourcolor}58.27\dd{1.37} & \cellcolor{ourcolor}70.07\dd{0.96} \\
\midrule
\multirow{2}{*}{Qwen3-4B} & Base    & 65.20 & 74.20 & 50.20 & 79.60 & 73.80 & 68.60 \\
 & \cellcolor{ourcolor}\textbf{TTT-NTP} & \cellcolor{ourcolor}64.40\dd{0.80} & \cellcolor{ourcolor}\textbf{77.20}\du{3.00} & \cellcolor{ourcolor}49.20\dd{1.00} & \cellcolor{ourcolor}79.20\dd{0.40} & \cellcolor{ourcolor}73.00\dd{0.80} & \cellcolor{ourcolor}68.60 \\
\midrule
\multirow{2}{*}{Qwen3-0.6B} & Base    & 53.53 & 57.87 & 38.82 & 70.13 & 50.24 & 54.12 \\
 & \cellcolor{ourcolor}\textbf{TTT-NTP} & \cellcolor{ourcolor}\textbf{54.03}\du{0.50} & \cellcolor{ourcolor}\textbf{59.72}\du{1.85} & \cellcolor{ourcolor}38.48\dd{0.34} & \cellcolor{ourcolor}70.13 & \cellcolor{ourcolor}\textbf{50.56}\du{0.32} & \cellcolor{ourcolor}\textbf{54.59}\du{0.47} \\
\bottomrule
\end{tabular}
\caption{General-capability evaluation comparing each released backbone (Base) to its TTT-NTP-trained backbone on standard commonsense and knowledge benchmarks (lm-evaluation-harness). Subscript arrows give the per-benchmark change of TTT-NTP relative to Base ({\textcolor{deltaup}{$\uparrow$}}~improvement, {\textcolor{deltadown}{$\downarrow$}}~decline); TTT-NTP values that beat Base are in \textbf{bold}. Aggregate scores are essentially unchanged across all four backbones.}
\label{tab:general}
\end{table}

\textbf{Long-context gains do not cost general capability.} Across all four backbones the aggregate over HellaSwag, ARC, PIQA, and MMLU shifts by at most about one point after TTT-NTP, and on three of the four it is flat or slightly \emph{higher}: $+0.47$ on Qwen3-0.6B, $+0.27$ on Llama-3.1-8B, and unchanged on Qwen3-4B. Per-task movements are similarly small---the largest is a few points on ARC-e---so the fast-weight write reshapes the long-context pathway without disturbing the model's knowledge and commonsense behaviour.

\textbf{The only notable regression is on Mistral.} Its aggregate slips by $0.96$, as a HellaSwag drop outweighs gains on ARC-e and PIQA. Even here the change is within roughly one point, so the long-context improvements of \cref{tab:ruler-main} come at no measurable general-capability cost across backbones.

\FloatBarrier

\section{Conclusion}
\label{sec:conclusion}

This work shows that long-context LLMs can benefit when test-time adaptation
is tied more directly to next-token prediction. TTT-NTP uses the observed
next token to train each adapted layer toward the next contextual state the
model already computes during a causal forward pass. Across four backbones
from three families spanning $0.6$--$8$B, it is the only adaptation that
consistently improves RULER Full-13 over the released model, with the gains
concentrated at the longest contexts and carrying over to real-world
long-document QA on LongBench-v2, while general capability is preserved. Two
controlled ablations locate the source of the gain: holding the training setup
fixed and changing only the value target
(\S\ref{sec:experiments:target-ablation}) shows the next-position target is what
helps, and varying only the inference-time solver
(\S\ref{sec:experiments:cf-ablation}) shows the gain is realized by a
regularized closed-form write whose key whitening, not the rank-one correlation
alone, is decisive. These results suggest that
better long-context use may come not only from larger windows, but from
adaptation signals that match how the model learned to predict text in the
first place.

\section*{Future Work}

Several directions follow naturally. Our study spans four backbones up to
$8$B and a single long-document corpus at a fixed token budget; scaling
TTT-NTP to larger checkpoints and more diverse pretraining mixtures, and
sweeping the data composition and compute, would test how far the next-position
signal carries. On the architecture side, our fast-weight write is developed for
full-attention decoders, and extending it to sliding-window and linear-attention
backbones---where the key statistics the closed-form solve relies on are local
rather than global---is an open and promising direction. Finally, the
closed-form inference-time write invites study in combination with native
long-context windows and retrieval-augmented pipelines, and broader evaluation
beyond retrieval (e.g.\ factuality and robustness) before deployment in
consequential settings.

\section*{Acknowledgment}

This research was supported in part by the National Science Foundation under Award \#IIS-2449768, the NVIDIA Academic Grant Program, and the Technical AI Safety Research Program at Coefficient Giving. The views and conclusions expressed in this work are those of the authors and should not be interpreted as representing the official policies or endorsements of the U.S. Government, the National Science Foundation, NVIDIA, or Coefficient Giving.

\bibliographystyle{colm2026_conference}
\bibliography{custom}

\appendix
\section{Training Hyperparameters}
\label{sec:appendix:training}

CPT, the In-Place TTT baseline, and TTT-NTP share the continual-pretraining recipe in \cref{tab:hp-train}; the per-backbone token budget is the only setting that varies across models. Each batch element is a single document, truncated to $32{,}768$ tokens, with no packing of documents; the outer loss is standard next-token cross-entropy. The Qwen3-0.6B checkpoint is a shared step-$100$ snapshot used as a smaller-scale consistency check rather than a compute-scaling study.

\begin{table}[h]
  \centering
  \begin{tabularx}{\linewidth}{@{}l X@{}}
    \toprule
    Setting & Value \\
    \midrule
    Corpus & Long-Data-Collections, pretraining split without the RedPajama-Book portion, i.e.\ RedPajama-ArXiv, a RedPajama sample, UL2 Oscar, Natural Instructions, P3, and a Pile subsample; mirrored on Hugging Face as emozilla/Long-Data-Collections-Pretrain-Without-Books\\
    Sequence length & $32{,}768$ \\
    Global batch size & $64$ \\
    Optimizer & AdamW, weight decay $0.1$ \\
    Peak learning rate & $5\times10^{-6}$, $5\%$ linear warm-up, cosine decay to $0$ \\
    Precision / sharding & bf16, FSDP \\
    Token budget & $0.4$B (Llama-3.1-8B), $0.1$B (Mistral-7B-v0.3), $2$B (Qwen3-4B), $0.2$B (Qwen3-0.6B) \\
    \bottomrule
  \end{tabularx}
  \caption{Continual-pretraining hyperparameters, shared by CPT, In-Place TTT, and TTT-NTP.}
  \label{tab:hp-train}
\end{table}

\section{Fast-Weight TTT Hyperparameters}
\label{sec:appendix:ttt}

TTT-NTP reuses each adapted layer's MLP down-projection as the fast weight, $\bm{W}_{\ell}=\bm{W}_{\ell}^{\mathrm{down}}$, with the current gated activation $\bm{z}_{\ell,t}$ as the key (no extra key projection) and the next-position state $\bm{h}_{\ell,t+1}$, passed through the learned $d{\times}d$ interface $\bm{W}_{\ell}^{\mathrm{proj}}$, as the value target. Under the inner-product inner loss (\cref{eq:ttt-update,eq:single-write}), $\eta$ scales the rank-one write directly rather than acting as a gradient step, so its tuned value varies widely across backbones (\cref{tab:hp-ttt}). On Qwen3-4B and Llama-3.1-8B, each chunk write $\Delta\bm{W}_{\ell}^{(c)}$ of \cref{eq:chunk-write} is divided by its spectral norm before entering the prefix sum of \cref{eq:prefix-weight}, which bounds the drift of the fast weight at large $\eta$ over these longer runs; Mistral-7B-v0.3 and Qwen3-0.6B use the raw write.

Writes are accumulated chunk-parallel (chunks of $1024$; an exclusive prefix sum lets chunk $c$ see only earlier chunks; \cref{sec:method:chunk}). $\bm{W}_{\ell}^{\mathrm{proj}}$ is initialized as a diagonal matrix whose diagonal is drawn from $\mathcal{N}(0,\,0.02^{2})$ and shared across the adapted layers (off-diagonal entries zero; Llama-3.1-8B instead uses a dense $\mathcal{N}(0,\,0.02^{2}/d)$ initialization) and trained jointly under the standard next-token CPT loss---the only TTT-NTP-specific parameter, adding a negligible $|\mathcal{A}|\,d^2$ weights.

At inference, the closed-form write of \cref{sec:method:cf} uses one fixed configuration for all backbones and context lengths: ridge regularizer $\lambda=1.0$, base step size $\eta=0.1$ with a per-layer cap $\|\eta_\ell\Delta\bm{W}_\ell\|_F/\|\bm{W}_\ell\|_F\le 0.1$, the regression fit on the last $8{,}192$ prefill tokens, the rows of the patched down-projection renormalized to their pretrained norms after the write, and the prompt key--value cache reused so that the write affects only decode-time computation.

\begin{table}[h]
  \centering
  \begin{tabular}{lcc}
    \toprule
    Backbone & TTT layers & Inner-loop $\eta$ \\
    \midrule
    Llama-3.1-8B ($32$ layers) & $\{0,6,12,18,24,30\}$ & $0.3$ \\
    Mistral-7B-v0.3 ($32$ layers) & $\{0,6,12,18,24,30\}$ & $0.15$ \\
    Qwen3-4B ($36$ layers) & $\{0,6,12,18,24,30\}$ & $3.0$ \\
    Qwen3-0.6B ($28$ layers) & $\{0,4,8,12,16,20,24\}$ & $2.6$ \\
    \bottomrule
  \end{tabular}
  \caption{Per-backbone fast-weight placement and inner-loop learning rate $\eta$ (chunk size $1024$ throughout).}
  \label{tab:hp-ttt}
\end{table}

\end{document}